\pdfoutput=1 %
\documentclass[letterpaper, 10 pt, conference]{ieeeconf}

\IEEEoverridecommandlockouts                               %

\usepackage{amsmath,amsfonts}
\allowdisplaybreaks
\usepackage{algorithmic}
\usepackage{algorithm}
\usepackage{array}
\usepackage[caption=false,font=footnotesize]{subfig}
\usepackage{textcomp}
\usepackage{stfloats}
\usepackage{url}
\usepackage{verbatim}
\usepackage{graphicx}
\usepackage{siunitx}
\usepackage{hyperref}
\hypersetup{pdftitle={Multi-View Latent Action Pretraining for Cross-Embodiment Robot Learning from Videos},pdfauthor={Oliver Hausd\"orfer, Felix Minzenmay, Mani Majd, Maximilian P. Christof, Shiyuan Zhang, Angela P. Schoellig}}
\usepackage{tikz}
\usepackage{placeins}

\usepackage[
 doi=false,isbn=false,url=false,eprint=false,
 backend=biber,
 style=ieee,
 bibstyle=ieee,
 citestyle=numeric-comp,
 sortcites=true,
 giveninits=true,
 maxcitenames=1,
 backref=false]{biblatex}
 \appto{\bibsetup}{\sloppy}
\AtEveryBibitem{%
  \clearfield{volume}%
  \clearfield{number}%
  \clearfield{pages}%
}
\DeclareRobustCommand{\citet}[1]{\citeauthor{#1}~\cite{#1}}

\usepackage{trimclip}
\usepackage{adjustbox}
\usepackage{threeparttable} 
\usepackage{booktabs}
\usepackage{multirow}
\usepackage{array}
\usepackage{makecell}
\usepackage{caption}
\usepackage{tikz}
\usepackage{tabularx}
\usepackage[caption=false,font=footnotesize]{subfig}
\usepackage{tabularx}
\newcolumntype{Y}{>{\centering\arraybackslash}X}

\usepackage[dvipsnames]{xcolor}
\usepackage{bbm}

\let\oldsubsection\subsection
\renewcommand{\subsection}{\vspace{-3pt}\oldsubsection}

\newcommand{\ieeetabstyle}{%
  \footnotesize
  \setlength{\tabcolsep}{0pt}%
  \renewcommand{\arraystretch}{1.15}%
}

\newcommand{\appendixrefname}{App.}
\newcommand{\appendixrefnamepl}{Apps.}
\DeclareRobustCommand{\appendixref}[1]{%
  \hyperref[#1]{\appendixrefname\nobreakspace\ref*{#1}}}
\DeclareRobustCommand{\appendixrefs}[2]{%
  \appendixrefnamepl\nobreakspace\hyperref[#1]{\ref*{#1}}~and~\hyperref[#2]{\ref*{#2}}}

\begin{document}

\title{Multi-View Latent Action Pretraining for\\Cross-Embodiment Robot Learning from Videos}

\author{Oliver Hausd\"orfer$^{1*}$, Felix Minzenmay$^{1*}$, Mani Majd$^{2\dag}$,\\
Maximilian P. Christof$^{1\dag}$, Shiyuan Zhang$^{1}$, and Angela P. Schoellig$^{1}$%
\thanks{$^{*}$Equal contribution. $^{\dag}$Equal contribution. $^{1}$The authors are with the Learning Systems and Robotics Lab, Technical University of Munich; and Munich Institute of Robotics and Machine Intelligence. This work was supported by the Robotics Institute Germany under BMBF grant 16ME0997K and by the Humboldt Professorship for Robotics and Artificial Intelligence. $^{2}$M.~Majd is with the University of Toronto. Research conducted while at LSY, TUM. Email: \href{mailto:oliver.hausdoerfer@tum.de}{oliver.hausdoerfer@tum.de}.}%
}

\maketitle

\begin{abstract}
Action-free videos are an appealing substitute for costly teleoperated data. Latent action models (LAMs) turn videos into a supervision signal by inferring abstract actions between frames. Existing LAMs observe a single camera view, which resolves 6D robot actions only ambiguously missing out-of-plane movements and potential occlusions. We propose a LAM that ingests synchronized multi-camera views using cross-attention to resolve depth and occlusions before compressing them into latent actions. The latent actions are fed into a policy in which a latent expert and an action expert are coupled by sequential flow matching around a frozen VLM backbone. We further evaluate in strict \textit{cross-embodiment} settings where task-specific demonstrations come from other embodiments. We show that dual-view LAM pretraining improves the decodability of ground-truth 6D actions by 17\% compared to single-view, while scaling single-view data by two orders of magnitude is not sufficient to fully recover out-of-plane motions. The downstream policy improves success rate by $+13.5\%$ compared to $+6.2\%$ for single-view LAM. A single cross-embodiment is sufficient for skill transfer, and adding a second viewpoint outperforms adding up to four further embodiments. The results transfer to hardware where demonstrations come from a human hand improving success rate by $+20\%$ over action-only policies in $850$ rollouts. We show that video and robot data are complementary, and that benefits of learning from video are task-dependent. Qualitative results are in the suppl. video. Website: {\urlstyle{same}\url{https://learnsyslab.github.io/multiview_latent_action_pretraining/}}.
\end{abstract}

\section{Introduction}
Data availability remains a limiting factor for learning robot policies. Recent works propose leveraging action-free video data that contains semantically rich information, which can aid robots in long-term task planning. One
approach infers latent actions from videos using an inverse dynamics model
(IDM), and conditions a downstream policy on
those latents~\cite{schmidt2024learningactactions,bruce2024geniegenerativeinteractiveenvironments,ye2025latentactionpretrainingvideos}. The latent actions are an abstract representation describing the transition dynamics in the video. Downstream policy success highly depends on how well the latent actions predict real robot actions. A number of approaches have been proposed to improve this transferability, including grounding latent actions in real robot actions or states\cite{chen2025villaxenhancinglatentaction,liang2025clamcontinuouslatentaction} and visual pre-processing approaches~\cite{nikulin2025latentactionlearningrequires,bu2026laofrobustlatentaction,bu2025univlalearningacttaskcentric}. However, we show that existing single-view approaches cannot sufficiently recover the full 6D poses that define robot actions. Single views capture actions only ambiguously: out-of-plane motion is underdetermined and parts of the scene may be occluded.

We therefore propose a Latent Action Model (LAM) that ingests synchronized multi-camera
views and attends to image patches across viewpoints---recovering depth and occlusion a single camera leaves ambiguous---before compressing them into a latent action. We demonstrate that this spatial awareness leads to better recoverability of 6D robot actions compared to single-view approaches. Crucially, multiple views are only required during LAM pretraining to improve latent action quality, but not during the policy learning stage where only a single external camera and wrist view are used.

A key advantage of learning from video is the ability to leverage cross-embodiment demonstrations, with the ultimate goal of learning from human demonstrations. Most existing works use only videos from the target robot embodiment, or leverage large-scale public datasets that confound different embodiments, tasks, and visual diversity~\cite{ye2025latentactionpretrainingvideos,chen2025villaxenhancinglatentaction,bu2025univlalearningacttaskcentric}, making performance attribution to cross-embodiment transfer difficult. We specifically evaluate our policy under \textit{cross-embodiment} demonstrations, where task-specific demonstrations come from other robots in simulation or humans on hardware. We then add a small number of demonstrations on the target embodiment and investigate performance as we scale. We find that a single cross-embodiment video source already enables transfer learning both in simulation and on hardware. Adding a second viewpoint consistently outperforms adding cross-embodiments, establishing multiple viewpoints as an affordable lever for policy performance. 

We propose to combine our LAM with a lightweight policy training jointly on the small action-labeled dataset and the much larger video-only set using sequential flow-matching experts. The sequential design provides a strong inductive prior, enabling the action expert to effectively leverage the latent plan. We benchmark our method on LIBERO~\cite{liu2023libero}, LIBERO-Plus~\cite{fei25libero-plus}, and MimicGen~\cite{mandlekar2023mimicgen}, and a real robot using 850 rollouts. \textbf{Our main contributions are:}
\begin{itemize}
    \item We propose an approach for multi-view IDM-style video pretraining and a downstream policy improving action-only learning by $+13.5\%$ in simulation and $+20\%$ on hardware.
    \item Training LAMs on in-domain multi-view videos raises the decodability of ground-truth actions by $17\%$ and policy success rate by $+7.3\%$ over single-view LAMs, and outperforms large-scale single-view pretraining, which does not fully recover out-of-plane motions.
    \item We show that a single cross-embodiment already enables skill transfer, and that videos from a second viewpoint outperform adding embodiments.
    \item On hardware, dual-view pretraining improves policy success rate by $+20\%$, and we demonstrate skill transfer from human-only demonstrations for novel tasks.
\end{itemize}

\section{Related Work}
\textbf{Latent Action Pretraining.} Inverse-dynamics models (IDMs) for latent action learning started with LAPO~\cite{schmidt2024learningactactions} and Genie~\cite{bruce2024geniegenerativeinteractiveenvironments}. The LAPO design of coupling an IDM and FDM (forward dynamics model) forms the template of modern LAM approaches. LAPA~\cite{ye2025latentactionpretrainingvideos} adopted this for robotics by finetuning a VLM on latent actions, and subsequently adding a small MLP head to predict actions. villa-X~\cite{chen2025villaxenhancinglatentaction} uses robots' proprioceptive states and CLAM~\cite{liang2025clamcontinuouslatentaction} and LAOM~\cite{nikulin2025latentactionlearningrequires} robot actions during IDM training for physical grounding of the latent actions. However, these methods ground latent actions solely in robot data from the target platform and leave human videos entirely ungrounded.
LAOF~\cite{bu2026laofrobustlatentaction} introduces optical flow as an intermediate representation to learn latent actions from videos, and UniVLA~\cite{bu2025univlalearningacttaskcentric} proposes using DINO features instead of pixels. Those methods still rely on video pre-processing from a single viewpoint, whereas we propose using multiple viewpoints to improve action recoverability.

\textbf{Multi-View Learning.} Most robot policies operate on a single external viewpoint~\cite{black2026pi0visionlanguageactionflowmodel,octomodelteam2024octoopensourcegeneralistrobot}, although recent work shows that 3D spatial perception yields stronger manipulation capabilities~\cite{ze20243ddiffusionpolicy,qian20243dmvp}. One approach injects explicit geometric reconstructions using point clouds~\cite{ze20243ddiffusionpolicy}, depth maps~\cite{liu2024robouniviewvisuallanguagemodelunified}, or voxels. These approaches rely on auxiliary depth sensors or complex reconstruction pipelines that introduce noise. %
Geometric structure can be learned implicitly using multi-view supervision: RoboUniView~\cite{liu2024robouniviewvisuallanguagemodelunified} pretrains a unified view representation from multi-view images, 3D-MVP~\cite{qian20243dmvp} pretrains the visual encoder via masked autoencoding over multi-view RGB-D renderings, and CLAMP~\cite{liu2026clamp} learns visual representations through a contrastive objective. MV-MWM~\cite{seo2023multiviewmaskedworldmodels} and ReViWo~\cite{pang2025reviwo} train multi-view encoder-decoders to build viewpoint-robust world models. Concurrent work MVP-LAM~\cite{lee2026mvplam} proposes latent action alignment between multiple viewpoints.

As latent action models aim to encode robot actions in 3D space, they stand to benefit substantially from 3D-aware pretraining. We propose to leverage multi-view pretraining of LAPA-style models to improve latent action learning.

\textbf{Cross-Embodiment Learning.} Videos are a natural source for cross-embodiment learning. Demonstrations from different robots can be architecturally absorbed through per-embodiment
input tokenizers~\cite{octomodelteam2024octoopensourcegeneralistrobot},
separate action heads~\cite{black2026pi0visionlanguageactionflowmodel},
or embodiment identity tokens~\cite{nvidia2025gr00tn1openfoundation} around a shared trunk. %
The \emph{visual} gap can be closed directly by editing pixels through inpainting at test time~\cite{chen2024miragecrossembodiment} or train time~\cite{chen2024roviaugrobotviewpoint}. The same method can bridge the human--robot gap by rendering a robot at
the estimated hand pose~\cite{lepert2025phantomtrainingrobots}. %
These methods make the mechanism of transfer interpretable, but depend on URDFs, renderers, or hand-pose estimators. XSkill~\cite{xu2023xskillcrossembodimentskill} discovers
embodiment-agnostic skills from unlabeled videos. UniAct~\cite{zheng2025universalactions} decodes a vector-quantized codebook of universal
behaviors, and \cite{bauer2025latentactiondiffusion}
contrastively aligns end-effector actions across hands and grippers. 

LAMs can achieve embodiment transfer from video alone without explicit reconstruction. The transfer question, however, remains confounded. Work that leverages aggregated datasets~\cite{chen2025villaxenhancinglatentaction} such as OXE~\cite{embodimentcollaboration2025openxembodimentroboticlearning}
varies embodiment, task, scene, and visual diversity all at once. Those settings make it difficult to quantify cross-embodiment skill transfer. We adopt a controlled
protocol in which task-specific video demonstrations stem from other robots in simulation and from human demonstrations on hardware.

\section{Method}\label{sec:method}

We consider language-conditioned visual manipulation. Given a
visual observation $o_t\in\mathcal{O}$ and a task instruction $\ell$, the policy acts in the robot's
action space $\mathcal{A}$. We seek a policy $\pi_\theta$ that predicts an action chunk $a_{t:t+K}$ of
$K$ future actions $a_t\in\mathcal{A}$ from the current context $x_t=(o_t,\ell)$. We learn this policy by imitation. The imitation objective for learning from a dataset
$\mathcal{D}_a=\{(\ell^{(j)},\,o^{(j)}_{0:T_j},\,a^{(j)}_{0:T_j})\}_j$ of action-labeled trajectories
minimizes the behavior cloning loss
\[
\min_\theta\ \mathbb{E}_{(x_t,\,a_{t:t+K})\sim\mathcal{D}_a}\big[\mathcal{L}_{\mathrm{BC}}\big(\pi_\theta(x_t),\,a_{t:t+K}\big)\big].
\]
The action labels are scarce and $\mathcal{D}_a$ holds only a small number of demonstrations. The second action-free dataset
$\mathcal{D}_v=\{(\ell^{(k)},\,o^{(k)}_{0:T_k})\}_k$ holds a much larger set of video-only demonstrations
of cross-embodiments. The problem we address is learning a policy $\pi_\theta$ that
leverages both datasets. 

Our method has two stages. A latent action
model (LAM) is pretrained on videos to infer a
\emph{latent action} for each observed transition (\autoref{subsec:lam}), using multiple viewpoints to provide spatial grounding (\autoref{subsec:multiview}). Then, a policy is trained on both sources jointly in a sequential manner (\autoref{subsec:policy}).

\subsection{Latent Action Model (LAM)}\label{subsec:lam}
Following \cite{schmidt2024learningactactions,bruce2024geniegenerativeinteractiveenvironments,ye2025latentactionpretrainingvideos}, our LAM couples an inverse dynamics model (IDM) that infers an abstract \textit{latent action} from
observed transitions with a forward dynamics model (FDM) that reconstructs the future from the present and that latent.

Unlike the common formulation, which infers the latent from a current and a future frame of a
\emph{single} viewpoint~\cite{ye2025latentactionpretrainingvideos,schmidt2024learningactactions}, our IDM ingests a synchronized
\emph{multi-view} frame pair $O_t=(o_t,\,o_{t+H})$ at a fixed frame offset $H$ --- the horizon the
policy plans over --- and encodes it jointly with a multi-view spatiotemporal
encoder $\Phi$, producing a fused token map for each of the two
frames, $f_t=\Phi_0(O_t)$ and $f_{t+H}=\Phi_H(O_t)$. A small bottleneck $B$ compresses each map. The
latent action is the difference of the bottleneck codes of the future and the current frame:
\begin{equation}
z_t \;=\; B\big(\Phi_H(O_t)\big) \;-\; B\big(\Phi_0(O_t)\big). \label{eq:latent}
\end{equation}
The FDM reconstructs the future frame $o_{t+H}$ from the current-frame tokens $e_t=E(o_t)$ --- the patch embedding $E$ shared with the encoder $\Phi$ --- and
the latent $z_t$, trained with a robust reconstruction loss $\rho$ (Huber loss):
\begin{equation}
\mathcal{L}_{\mathrm{LAM}} \;=\; \mathbb{E}_t\, \rho\Big( \operatorname{FDM}\big(\operatorname{sg}[e_t],\, z_t\big),\; o_{t+H}\Big). \label{eq:fdm}
\end{equation}
The stop-gradient $\operatorname{sg}[\cdot]$ on the current-frame tokens forces the only
trainable path that lowers future-reconstruction error to run \emph{through the latent} (and not through $E$), so $z$
captures the controllable change rather than static scene content. 

\begin{figure*}[t] 
    \vspace{2mm}
    \centering
    \includegraphics[width=\textwidth]{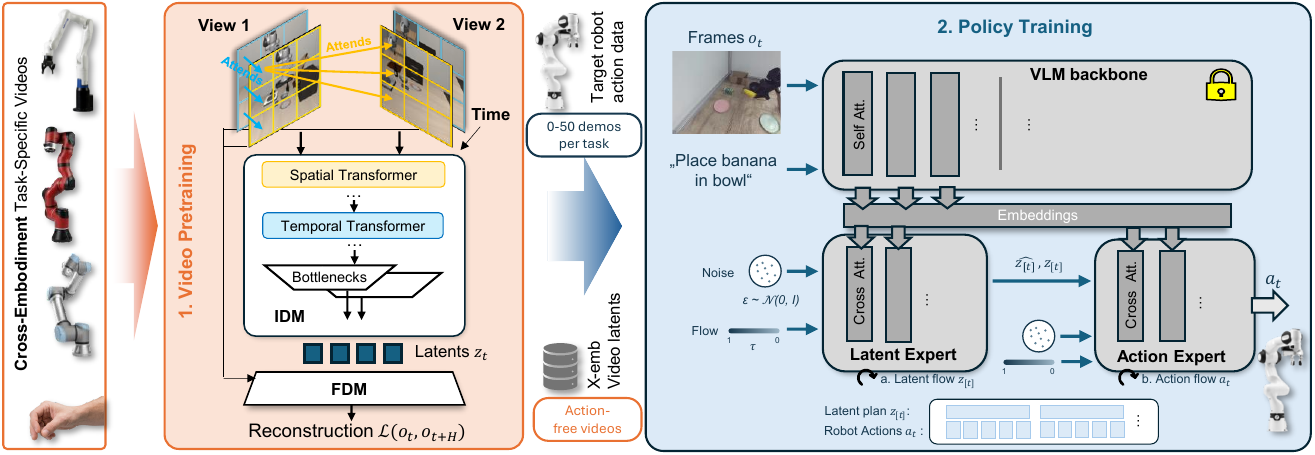} %
    \caption{\textbf{Method}: We combine latent action pretraining from in-domain cross-embodiment videos with a downstream flow-matching policy. First, latent actions are inferred from a large cross-embodiment video dataset. Second, a sequential policy is learned on the latent actions and a small set of demonstrations on the target embodiment. The LAM uses a spatio-temporal transformer to merge multiple viewpoints before computing the latents. The spatial block attends to patches of other viewpoints, yielding latent actions grounded in 3D space. Latent action labels and real robot actions are jointly used to train a hierarchical downstream policy. The policy uses sequential flow matching experts and outputs actions applied to the robot.}
    \label{fig:wide_figure}
\end{figure*}

\subsection{Multi-Viewpoint Fusion}\label{subsec:multiview}
An action manifests as 6D poses of the end-effector, which a single RGB view
captures only ambiguously --- from any single viewpoint out-of-plane motion is
underdetermined, and parts of the scene may be occluded.
We hypothesize that multiple synchronized viewpoints give the LAM a better 3D understanding of
the scene and hence more spatially-grounded latents than a single view.

\textbf{Multi-view spatiotemporal encoder.} Each RGB frame $I^{(v)}_t$ (view $v\in\{1,\dots,V\}$,
window offset $t\in\{0,H\}$) is split into $N$ patches; patch $p$ of view $v$ is embedded by the shared patch embedding $E$
and tagged with a learned camera-identity embedding $c_v$, giving tokens $\mathbf{x}_{t,v,p}$.
These pass through a stack of spatial attention blocks followed by a stack of temporal attention
blocks. We borrow the factorized space-time attention that has proven effective for video understanding in computer vision \cite{arnab2021vivit} and adapt it to the multi-view setting, similar to recent world models \cite{bruce2024geniegenerativeinteractiveenvironments,seo2023multiviewmaskedworldmodels}. Every block uses cosine multi-head attention~\cite{liu2022swinv2}, where queries and keys
are $\ell_2$-normalized, rescaled by learned per-dimension scales, and multiplied by a
fixed temperature before the softmax. Rotary position encodings
(RoPE)%
 are applied to the normalized queries and keys.

\textbf{Joint cross-view attention.} We let the spatial block attend over the union of all views' patches — so a patch in one view attends to patches in every other view for a given timestep $t$. This way, attention can resolve geometry and occlusion across viewpoints. The temporal block then attends across the two frames at each fixed location. Views are pooled into the fused map that defines the latent:
\begin{align}
\text{cross-view spatial:}\ \ & \mathbf{x}_{t,\cdot,\cdot}\ \leftarrow\ \operatorname{Attn}\big(\{\mathbf{x}_{t,v,p}\}_{v\in[V],\,p\in[N]}\big), \notag\\*
& \text{2-D RoPE on } p; \notag\\[2pt]
\text{temporal:}\ \ & \mathbf{x}_{\cdot,v,p}\ \leftarrow\ \operatorname{Attn}\big(\{\mathbf{x}_{t,v,p}\}_{t\in\{0,H\}}\big), \label{eq:crossview}\\*
& \text{1-D RoPE on } t; \notag\\[2pt]
\text{view fusion:}\ \ & f_{t}\ =\ \tfrac{1}{V}\textstyle\sum_{v=1}^{V}\mathbf{x}_{t,v,\cdot}. \notag
\end{align}
Encoding the views \emph{jointly} --- rather than with independent per-view encoders and late fusion ---
lets the latent exploit cross-view correspondence at every layer. This factorization suits our setting: because the viewpoints are
synchronized and static, patches across views correspond to shared 3D structure at each instant,
so cross-view attention in the \emph{spatial} stage resolves per-frame geometry (depth, occlusion)
while the \emph{temporal} stage isolates how that geometry changes via the action. The resulting maps
$f_t=\Phi_0(O_t)$ and $f_{t+H}=\Phi_H(O_t)$ feed the difference latent of ~\autoref{eq:latent}.

\subsection{Latent-Augmented Policy}\label{subsec:policy}
The latent plan, as well as robot actions, are jointly consumed by the policy. We design a hierarchical policy that first predicts a latent plan from context, then an action chunk
conditioned on that plan, factorizing the joint distribution as
\begin{equation}
p_\theta\big(a_{t:t+K},\, z_{[t]} \mid x_t\big) \;=\; \underbrace{p_\theta\big(z_{[t]}\mid x_t\big)}_{\text{latent plan}} \cdot \underbrace{p_\theta\big(a_{t:t+K}\mid x_t, z_{[t]}\big)}_{\text{action chunk}}, \label{eq:factor}
\end{equation}
where $x_t=(o_t,\ell)$ is the context (\autoref{fig:wide_figure}). Mirroring the action chunk of $K$ actions, the latent plan
$z_{[t]}=(z_{t+jH})_{j=0}^{M-1}$ is a chunk of $M=K/H$ latents spanning the same interval, but with fewer
entries since each latent covers the $H$ frames of the LAM's frame offset
(\autoref{eq:latent}). This differs from joint latent--action models
\cite{chen2025villaxenhancinglatentaction} where the action depends on the plan through a joint denoising target instead of an explicit conditioning interface.

The interaction between the backbone and action experts alternates cross-attention to the backbone's keys and values with self-attention over the expert's own tokens, similar to SmolVLA \cite{shukor2025smolvla}. Each expert is
trained by conditional flow matching: for a target $y$ (a latent plan or action chunk) with
conditioning $c$, noise $\varepsilon\sim\mathcal{N}(0,I)$ and time $\tau$,
\begin{equation}
\begin{aligned}
x^\tau &= \tau\varepsilon+(1-\tau)y,\quad u=\varepsilon-y,\\
\mathcal{L}_{\mathrm{fm}}(y,c) &= \mathbb{E}_{\tau,\varepsilon}\big\|v_\theta(x^\tau,\tau,c)-u\big\|^2,
\end{aligned}
\label{eq:fm}
\end{equation}
with denoised estimate $\hat y = x^\tau-\tau\,v_\theta(x^\tau,\tau,c)$. The two experts differ only in their
conditioning: $c_z=x_t$ for the latent expert and
$c_a=(x_t,\,z_{[t]})$ for the action expert.

Every sample of $\mathcal{D}_v$ and
$\mathcal{D}_a$ carries latent labels, while only $\mathcal{D}_a$ carries action labels. We formulate a loss
that jointly trains a policy on both targets:
\begin{equation}
\mathcal{L} \;=\; \lambda_a\,\mathbb{E}_{i\sim\mathcal{D}_a}\big[\ell_a^i\big] \;+\; \lambda_z\,\mathbb{E}_{i\sim\mathcal{D}_v\cup\mathcal{D}_a}\big[\ell_z^i\big], \label{eq:loss}
\end{equation}
where $\ell_a^i,\ell_z^i$ are the per-sample flow-matching losses of ~\autoref{eq:fm} averaged over chunk steps and dimensions, and $\lambda_a,\lambda_z$ are weights. We train the action expert using scheduled sampling \cite{bengio2015scheduledsampling} to enable conditioning on both ground-truth plans $z_t$ and its own noisy predictions $\hat{z_t}$ during training.

\section{Implementation Details}\label{sec:impl}

\textbf{Implementation Details}. Our LAM trains from scratch only on in-domain videos on four NVIDIA
H100 GPUs for $12$\,h, and the flow-matching experts for $13$\,h on one. \autoref{tab:impl} lists the model and training parameters. 

\textbf{Latent Action Probing}.
We use \textit{probing}~\cite{nikulin2025latentactionlearningrequires,chen2025villaxenhancinglatentaction}---fitting a lightweight model to predict ground-truth robot actions from frozen latent embeddings---as a metric for evaluating the quality of the learned latent actions. We report the
  coefficient of determination,
  \begin{equation}
  R^2 = 1 - \frac{1}{D}\sum_{d=1}^{D} \frac{\sum_i (a_{i,d}-\hat a_{i,d})^2}{\sum_i (a_{i,d}-\bar a_d)^2}, \quad \hat a_i = f_\theta(z_i),
  \end{equation} 
  where $f_\theta$ is a small MLP probe fit on the frozen latents $z_i$, $i$ indexes heldout
  transitions in the dataset, and $D{=}7$ is the action dimension. The probe is small and the latents are frozen, so it cannot recover information that is absent in the latent. It reflects how well the latent predicts robot actions.

\textbf{Oracle Latents}. To \textit{estimate} an upper bound of policy performance assuming perfectly reconstructed latent actions, we introduce oracle latents. The oracle latents are a compression of the actions over the chosen time horizon $H$. The oracle latents are computed directly from the available ground-truth 7D actions $a_t$:
$$
z_t^{\text{oracle}} = \sum_{k=0}^{H-1} a_{t+k},
$$
and reflect the robot's actions over the latent horizon. We train the downstream policies identically to using the real latents. It serves as an estimation of the upper performance boundary we expect when adding video data to the policy. Remaining gaps are attributable to how well latent action learning actually helps task-solving, or how the policy can leverage latent actions.

\textbf{Simulation Benchmarks}. We benchmark on LIBERO~\cite{liu2023libero}, LIBERO-Plus~\cite{fei25libero-plus}, and MimicGen~\cite{mandlekar2023mimicgen}. In all suites, we generate the dataset with a second side-view camera for LAM training. The deployed embodiment is Franka Panda for all suites, and cross-embodiments are UR5e, KUKA LBR iiwa, Sawyer, and Kinova Gen3 (those only contribute videos). \textit{LIBERO} is a benchmark with four task suites: Spatial, Object, Goal, and LIBERO-10. Each task suite contains 10 tasks, and each task contains 50 demonstrations. We replay the source demonstrations on the different embodiments, yielding 160 action-annotated Panda and 4,293 video cross-embodiment demonstrations across 32 tasks. In addition, we define eight tasks (two per suite) as \textit{heldout tasks} that never see any robot action from Franka Panda to test generalization under video-only settings. LIBERO-90 adds a further 13,338 episodes (3,936 Panda) of off-task play data. \textit{LIBERO-plus~\cite{fei25libero-plus}} adds visual disturbances on top of LIBERO, and we follow the same generation recipe. \textit{MimicGen}~\cite{mandlekar2023mimicgen} is a data-generation scheme including a suite of 28 challenging, high-precision tasks. Retaining only task-embodiment combinations with $>20\%$ generation success rate yields 107 of 140 task$\times$embodiment combinations in 42.3k episodes, of which 12.7k are Panda. We use 22 tasks as play data and evaluate on 6 eval tasks.

\begin{table}[t]
    \vspace{3mm}
\caption{Model and training hyperparameters.}
\label{tab:impl}
\centering
\setlength{\tabcolsep}{3pt}
\renewcommand{\arraystretch}{1.0}
\footnotesize
\begin{tabular*}{\columnwidth}{@{\extracolsep{\fill}}lll@{}}
\toprule
\multicolumn{3}{@{}l}{\textbf{Latent action model} ($344$M params)}\\
\midrule
Hidden dim $d$ & \multicolumn{2}{l@{}}{$1024$} \\
Spatial / temporal / FDM layers & \multicolumn{2}{l@{}}{$8$ / $8$ / $8$ (FDM cross-attends $z$)} \\
Bottleneck $B$, latent & \multicolumn{2}{l@{}}{$1024{\rightarrow}8$ strided conv; $1{\times}8$ cont.} \\
Input res.\ / offset $H$ & \multicolumn{2}{l@{}}{$256{\times}256$ / $5$ (\SI{0.25}{\second} @ \SI{20}{\hertz})} \\
Reconstruction loss $\rho$ & \multicolumn{2}{l@{}}{Huber, $\beta{=}0.1$} \\
\midrule
\multicolumn{3}{@{}l}{\textbf{Policy} ($202$M trainable / $552$M total)}\\
\midrule
Backbone & \multicolumn{2}{l@{}}{SmolVLM2-500M (frozen)} \\
Hidden dim (backbone / expert) & \multicolumn{2}{l@{}}{$960$ / $720$} \\
Expert layers & \multicolumn{2}{l@{}}{$16$ (each expert)} \\
Chunk $K$ / plan $M$ & \multicolumn{2}{l@{}}{$50$ / $10$ (stride $H{=}5$)} \\
Loss weights / teacher forcing $p$ & \multicolumn{2}{l@{}}{$\lambda_a{=}\lambda_z{=}1$ / $1.0{\rightarrow}0.2$} \\
\midrule
\textbf{Training} & \textbf{LAM} & \textbf{Policy} \\
\midrule
Steps & $70$k & $100$k \\
Batch (per GPU $\times$ GPUs) & $32{\times}4$ & $80{\times}1$ \\
LR peak $\rightarrow$ final (cosine decay)& $10^{-4}\!\rightarrow\!10^{-6}$ & $10^{-4}\!\rightarrow\!2.5{\times}10^{-6}$ \\
Optimizer & AdamW, bf16 & AdamW, bf16 \\
\bottomrule
\end{tabular*}
\end{table}

\section{Evaluation}\label{sec:eval}

\subsection{Policy Performance under Video Supervision (\autoref{tab:main})}\label{subsec:policyperf}

We evaluate policies under settings with scarce action demonstrations on the deployed embodiment, and the in-domain cross-embodiment demonstrations always only contribute videos. For comparison, we train policies on the scarce action budget only without using video data achieving $42.7\%$ SR, and on the full action budget of the entire dataset achieving $75.1\%$. Those values provide upper and lower reference points. To verify that the lower reference is a data gap and not a model gap, we finetune a model with significantly more parameters than ours ($\pi_{0.5}$ $4.1$B vs. our $202$M) and $4\times$ more compute. $\pi_{0.5}$ does not recover the full action budget SR. These results show that the little action availability limits policy performance, and leaves room for improvement when adding video data.

\begin{table}[!t]
    \vspace{3mm}
\caption{Rollout Success Rate (\%) Across Benchmarks}
\label{tab:main}
\centering
\ieeetabstyle
\newcommand{\ind}{\hspace{0.8em}}
\newcommand{\vsep}{\hspace{1em}\vline\hspace{1em}}
\renewcommand{\T}{}  %
\newcommand{\B}{\rule[-1.2ex]{0pt}{0pt}}  %
\newcommand{\sect}[1]{\multicolumn{4}{@{}l@{\vsep}}{\T\emph{#1}} & \\}
\setlength{\aboverulesep}{0pt}
\setlength{\belowrulesep}{0pt}
\begin{threeparttable}
\begin{tabular*}{\columnwidth}{@{\extracolsep{\fill}}lrrr@{\extracolsep{0pt}\vsep}r@{}}
\toprule
\T\B Method & MimicGen & LIBERO & LIB.+ & Mean \\
\midrule
\sect{Full action budget}
\ind\B Ours & 65.3 & 75.0 & 85.1 & 75.1 \\
\midrule
\sect{Low action budget}
\ind Ours (202M params) & 25.7 & 52.6 & 49.8 & 42.7 \\
\ind\B $\pi_{0.5}$ (4.1B params) & 32.0 & 53.0 & 54.0 & 45.0 \\
\midrule
\sect{Low action budget (our policy) + LAM}
\ind villa-X (state grnd., VQ latent)$^{\dag}$ & 18.8 & 42.8 & 45.1 & 35.6 \\
\ind villa-X (state grnd., cont. latent)$^{\dag}$ & 22.0 & 43.8 & 51.5 & 39.1 \\
\ind \textbf{Ours (single-view)} & \textbf{31.0} & \textbf{53.5} & \textbf{62.1} & \textbf{48.9} \\
\ind MVP-LAM$^{\ddag}$ & 24.7 & 57.6 & 65.4 & 49.1 \\
\ind UniVLA (DINOv3 features)$^{\ddag}$ & 28.7 & 59.3 & 60.9 & 49.6 \\
\ind CLAM (action grounding)$^{\ddag}$ & 32.3 & 57.2 & 63.1 & 50.9 \\
\ind LAOF (flow constraint)$^{\ddag}$ & 33.4 & 58.8 & 60.5 & 50.9 \\
\ind LAPA-pretrained & 30.4 & 59.9 & \underline{67.7} & 52.7 \\
\ind \textbf{Ours (multi-view)} & \underline{\textbf{38.7}} & \underline{\textbf{62.3}} & \textbf{67.6} & \underline{\textbf{56.2}} \\
\ind\B Oracle$^{*}$ & 39.2 & 71.0 & 72.4 & 60.9 \\
\bottomrule
\end{tabular*}
\begin{tablenotes}[flushleft]\scriptsize
\item[] \textit{Low action budget}: five action-annotated demonstrations per task for Franka Panda on LIBERO and MimicGen, and one for LIBERO+.
$^{\dag}$Pretrained latent-action model from the authors' provided checkpoints. $^{\ddag}$Re-implemented on our architecture based on the
authors' released code, so that only the named component differs from ours.
$^{*}$ Oracle: see \autoref{sec:impl}. \underline{Best} is underlined.
\end{tablenotes}
\end{threeparttable}
\end{table}

Cross-embodiment video supervision in the single-view setting improves SR by an absolute $+6.2\%$ for our method. Multi-view pretraining improves policy SR by another $+7.3\%$, resulting in $+13.5\%$ over action-only. This demonstrates that multiple viewpoints add significant information to the latent that the policy can leverage to determine robot actions. The gains from adding multi-view are concentrated in better 6D information contained in the latent actions as we show in \autoref{fig:probing}. The improvements from adding a second viewpoint are more significant than baseline methods that propose visual preprocessing of single-view inputs (LAOF), action grounding (CLAM), or use large-scale pretraining (LAPA).

\begin{figure*}[b]
  \centering
  \includegraphics[width=\textwidth]{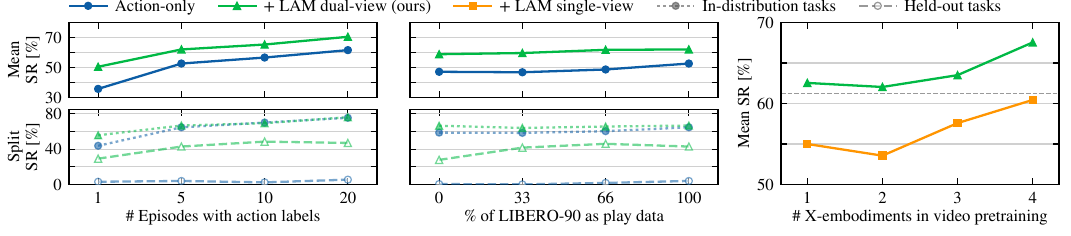}
  \caption{\textbf{Scaling with Data.} Policy success rate (SR) when scaling \textbf{(left)} the number of action-annotated demonstrations per task on the deployed embodiment Franka Panda, \textbf{(middle)} the fraction of LIBERO-90 used as off-task play data for latent video pretraining increasing task diversity. Top row shows mean success rate over all 40 tasks; bottom row split by \textit{in-distribution} and \textit{heldout} tasks (see \autoref{sec:impl}). Successes on heldout tasks prove transfer from cross-embodiment video demonstrations. \textbf{(Right)}: The number of cross-embodiments contained in the video dataset (1 = iiwa, 2 = +kinova3, 3 = +sawyer, 4 = +ur5e) for single- and dual-view pretraining. Dual-view pretraining outperforms adding up to four cross-embodiments.}
  \label{fig:libero_scaling_row}
\end{figure*}

MVP-LAM also pretrains on both views, yet significantly trails our approach. MVP-LAM couples viewpoints through a cross-view reconstruction loss but encodes them separately, whereas our LAM jointly encodes both viewpoints into a single latent that resolves depth. CLAM-style action-grounding improves policy success only by $+2.0\%$ compared to our $+7.3\%$. This is because action-grounding can only use a subset of the available dataset for grounding instead of the entire dataset---only the fraction of data carrying actions.

Training on a small dual-view dataset outperforms a baseline pretrained on orders of magnitude more single-view data. The closest comparison to ours is initializing our single-view variant pretrained weights from LAPA trained on 970K robot trajectories~\cite{ye2025latentactionpretrainingvideos}, and then finetuning on our data. Despite being pretrained on two orders of magnitude more data, it does not outperform our dual-view LAM. The advantage of LAPA-pretraining over no pretraining originates from general visual robustness, but not from fully recovered latent action information in the depth dimension (\autoref{fig:probing}). For instance, LAPA performs best on LIBERO+, which introduces visual disturbances. Villa-X was pretrained on $5.2$M human and robot trajectories, and we re-use their checkpoints without finetuning, as no code is available. The best version does not improve on baseline. More single-view data cannot fully resolve the depth ambiguity of a 2D projection. Thus, pretraining on orders of magnitude more single-view data does not necessarily substitute for a second, calibrated viewpoint.

\subsection{Scaling With Sources of Demonstrations (\autoref{fig:libero_scaling_row})}\label{subsec:scaling}

\autoref{fig:libero_scaling_row} (a) scales the available action-annotated demonstrations on the deployed embodiment, while keeping the rest fixed. For all budgets, video pretraining consistently improves on action only. The results are separated by in-distribution and heldout tasks. On \textit{in-distribution} tasks, where the target embodiment receives a varying number of action-annotated demonstrations, the advantage of video data decreases as more demonstrations are added. The \textit{heldout} tasks only receive cross-embodiment videos but no action-annotated demonstrations, and the SR remains near zero without adding video demonstrations (action-only). Adding cross-embodiment video demonstrations achieves $48.6\%$ SR on heldout-tasks. Thus, the video behavior of cross-embodiment demonstrations successfully transfers to the unseen tasks. We observe that increasing the demonstrations for other tasks also increases performance on heldout tasks. This is because LAM and policy have more data to ground latent actions in real actions.

In ~\autoref{fig:libero_scaling_row} (b) we vary the amount of play data, i.e., data from non-eval tasks. On \textit{in-distribution tasks}, the size of the play data has marginal impact, with 0 play data performing as well as the full play dataset of $\approx10\,\text{k}$ cross-embodiment and $3\,\text{k}$ panda trajectories ($66.6\%$ vs $66.8\%$). For \textit{heldout-tasks} that never receive action demonstrations, a larger off-task play data corpus improves SR, as the policy has been exposed to more diverse data and scenes, helping generalization. For generalization to novel tasks a large play dataset is desirable.

\begin{figure*}[t]
  \vspace{2mm}
  \centering
  \includegraphics[width=\textwidth]{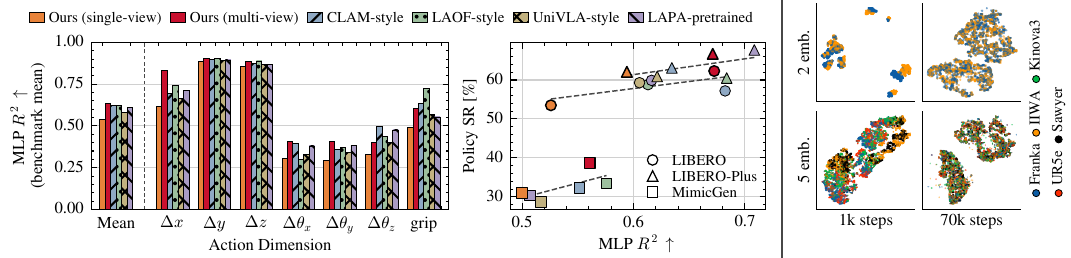}
  \caption{\textbf{Left: Latent Action Probing}. An MLP probe reconstructs ground-truth robot actions from the latents. Adding the second view improves decodability in the directions that are only ambiguously observed by single-view. \textbf{Right: Cross-embodiment Learning}. t-SNE showing 2D correspondence of latent actions to embodiments. Separation by color indicates separate latent per embodiment before training, while mixed colors show a latent space shared between embodiments. Shared latents already occur for a single cross-embodiment present in pretraining.}
  \label{fig:probing}
\end{figure*}

In~\autoref{fig:libero_scaling_row} (c), we scale the number of cross-embodiments that contribute videos for both single- and dual-view training. In general, adding a second viewpoint is a cheaper way to improve policy performance compared to adding additional embodiments. A single cross-embodiment with a dual-view setting outperforms four additional embodiments in single-view settings. For both single-view and dual-view, a single embodiment recovers most of the gains for the \textit{heldout tasks}. For instance, the dual-view LAM achieves $58.3\%$ with one cross-embodiment compared to $58.9\%$ with four (not shown in the figure). We also show behavioral transfer across embodiments in \autoref{fig:probing} (right) and \autoref{fig:realworld_probe}. After convergence, the latents of different embodiments merge into a single manifold, i.e., different robots reuse the same latents, and the latent encodes what the action is and not which robot executes it. Similarly, on hardware, we show that the predicted latents transfer between human and robot embodiments (\autoref{fig:realworld_probe}).

\subsection{Latent Quality for Dual-View LAMs and Predictiveness for Policy Success Rate (\autoref{fig:probing})}\label{subsec:latentquality}

The second view improves action decodability $R^2$ from $0.54$ for single-view to $0.63$ for dual view ($+17\%$). The gains concentrate in the dimensions $\Delta x$ ($0.62$ to $0.83$ avg.), and $\Delta r_y$ ($0.29$ to $0.41$ avg.), which are the out-of-plane motions for the first camera view. Thus, the second view effectively recovers action-decodability for 6D motions and does not merely overfit to existing dimensions. Cross-view attention resolves movements across views, so latents represent 3D motion rather than its ambiguous 2D projections into a single view.

In contrast, large-scale pretraining (LAPA) primarily improves reconstruction of existing in-plane motions and partially gripper actions, but does not recover motions in the out-of-plane direction. This shows that simply scaling single-view data collection is not an effective way to recover 3D motions. Instead, collecting more high-quality calibrated multi-view datasets for robot learning is desirable. Similarly, baselines that use visual preprocessing methods on single-view, such as optical flow constraints (LAOF) or DINO encoding (UniVLA) primarily improve on in-plane motions and gripper actions, but not in out-of-plane directions.

\begin{figure*}[!t]
    \vspace{2mm}
    \centering
    \includegraphics[width=\textwidth]{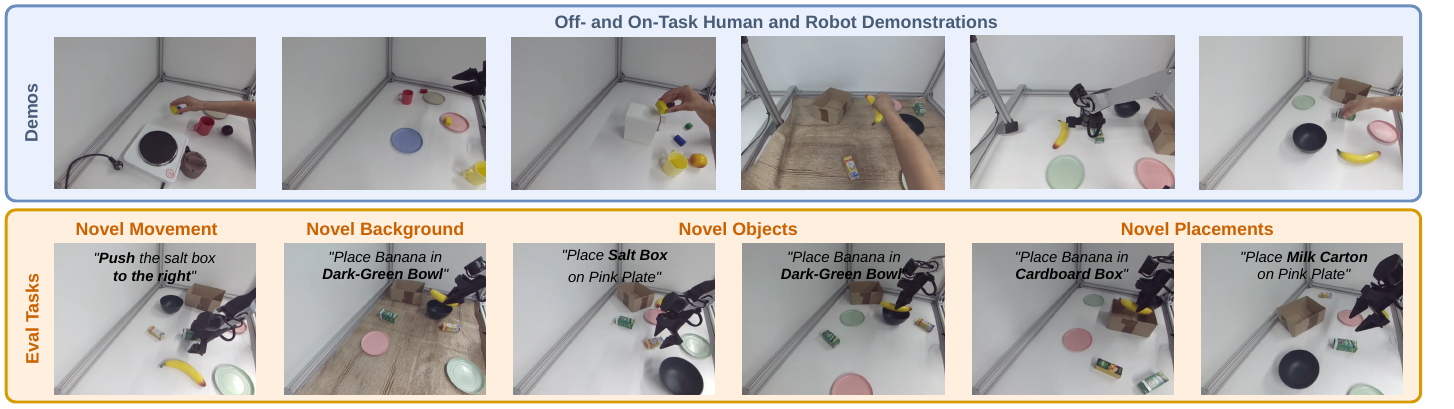}
    \caption{\textbf{Real World Tasks}: We evaluate the policies and skill-transfer on six tasks consisting of novel movements, novel background, novel objects, and novel placements. For each eval task, we collect 50 human demonstrations and evaluate under settings with 0--50 robot demonstrations per task. We additionally collect an off-task play dataset  across 32 tasks, totaling 1000 human demonstrations and 600 robot demonstrations.}
    \label{fig:real_world_tasks}
\end{figure*}

\begin{figure*}[!t]
  \centering
  \includegraphics[width=\textwidth]{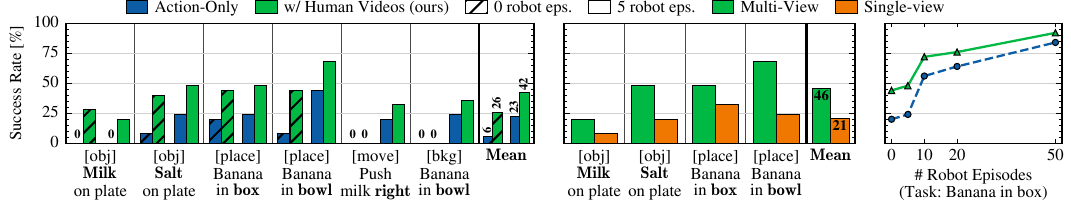}
  \caption{\textbf{Evaluation}: Each datapoint represents 25 rollouts.
  \textbf{Left:} We observe successes without using any robot data for new tasks. On average, using human demonstrations without any robot demonstrations increases SR by $+20\%$, and with 5 robot demonstrations by $+19\%$. Tasks with novel placement locations ([place]) are easiest to solve, followed by novel objects ([obj]). \textbf{Middle:} Multi-view video pretraining scores $46\%$ against $21\%$ for single-view significantly increasing SR.
  \textbf{Right:} Scaling of SR with increasing robot demonstrations for one task: video and robot demonstrations are complementary.}
  \label{fig:realworld_row}
\end{figure*}

For CLAM-style action grounding, we find that action decodability is separated by embodiments: $R^2=0.74$ on Panda vs. $0.57$ on the others. This is due to an asymmetric training scheme that leverages action grounding only on the target embodiment for which actions are available. This leads to a separated latent space and hinders cross-embodiment learning. For instance, action grounding improves policy SR by $+5.0\%$ on in-distribution tasks for which Franka Panda demonstrations are available, but not on heldout tasks that only see cross-embodiment videos ($-1.4\%$ SR). In comparison, for our dual-view pretraining action decodability stays flat across embodiments ($0.58$ Panda vs. $0.61$ others), and we observe significant gains even on heldout tasks ($+15.1\%$), proving effective skill transfer.

Reconstructing gripper actions from video is particularly difficult, as they are often low-displacement events that are weakly reflected in pixel changes. CLAM action grounding and LAOF flow constraint help here, as ground-truth gripper actions or informative flow loss help to capture those fine movements. Flow constraints and action grounding are both complementary to multi-view learning. We show that failed gripper actions are a major failure case when learning from video data (\autoref{subsec:realworld}).

The probing metric reliably identifies degenerate LAMs that do not capture the relative movement between frames. For instance, villa-X's quantized latents probe at only $0.26$  and yield the lowest downstream policy SR (not shown in the figure). Across the 18 LAM-suite combinations in \autoref{fig:probing}, probe $R^2$ and SR correlate with $r = 0.83$. Within a benchmark, LAMs whose $R^2$ lies within $0.05$ of each other are ordered correctly in $59\%$ of cases.

\subsection{Gap of Latent-Augmented to Action-Supervised Policies}\label{subsec:oracle}

The oracle latents replace the learned latents with targets computed from the ground-truth actions (\autoref{sec:impl}), \textit{estimating} perfect latent reconstruction. The ours-vs-oracle gap therefore isolates \emph{teacher quality}, i.e., how much does policy performance improve with perfect latent actions.

Our multi-view latents recover $99\%$ of the oracle baseline for MimicGen, $88\%$ for LIBERO, and $93\%$ for LIBERO+. Thus, our LAM is close to the performance expected under estimated perfect latent actions, highlighting its reconstruction quality.

The remaining gap between oracle and the \textit{full action} policies is larger on some benchmarks. Oracle recovers $95\%$ of the full action budget policies for LIBERO, $85\%$ for LIBERO+, and $60\%$ for MimicGen. The oracle latents reflect our LAM training settings; specifically, they \textit{compress} the ground-truth actions over the latent's horizon. Tasks that require high precision have limited benefit from sparsely annotated latent actions and require dense action chunks instead. The performance bottleneck is thus not necessarily latent reconstruction, but settings of the LAM and latent usefulness for the particular tasks. Analysis further shows that oracle latents lower overall failures, but do not recover grasping performance. Full action supervision cuts grasping failure to $42.5\%$. Grasping is therefore limited by action supervision, not by latent quality.

\begin{figure*}[t]
    \vspace{2mm}
  \centering
  \includegraphics[width=\textwidth]{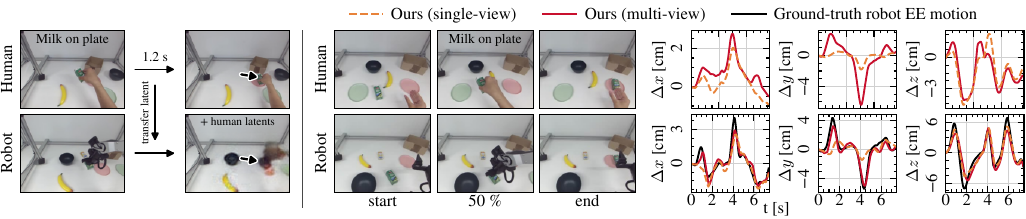}
  \caption{\textbf{Human to Robot Transfer. Left}: We apply the latent action computed on a human frame to a robot frame from an eval rollout not seen during training: both yield the same movement to the right. \textbf{Right}: Motion decoded from the predicted latents for a human and a robot video. Both motions decode into similar motions in robot end-effector space. The multi-view LAM shows significantly better reconstruction of the ground-truth robot motion in $y$-direction. See more qualitative results in the supplementary video.}
  \label{fig:realworld_probe}
\end{figure*}

\subsection{Real World Experiments (\autoref{fig:real_world_tasks}, \autoref{fig:realworld_row} \autoref{fig:realworld_probe})}\label{subsec:realworld}

On hardware, we use human demonstrations as video data. The policies run on a TRLC-DK1~\cite{trlc2026} robot.

\begin{figure}[t]
      \vspace{2mm}
  \centering
  \begin{minipage}[t]{0.62\columnwidth}
    \vspace{0pt}
    \includegraphics[width=\linewidth]{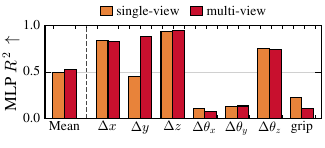}
  \end{minipage}\hfill
  \begin{minipage}[t]{0.35\columnwidth}
    \vspace{0pt}\raggedright
    \caption{\textbf{Probing on Hardware}. Per-dimension action probing for heldout episodes on hardware. The second view
    raises $R^2$ in $\Delta y$ from $0.45$ to $0.88$.}
    \label{fig:realworld_probe_bars}
  \end{minipage}

  \vspace{-2mm} %

  \subfloat[Missed grasp.]{%
    \includegraphics[width=0.32\columnwidth,height=0.31\columnwidth]{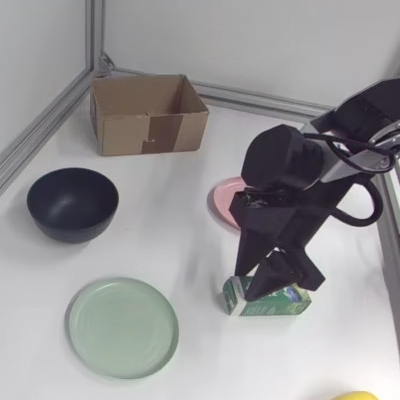}%
  }\hfill
  \subfloat[Wrong placement.]{%
    \includegraphics[width=0.31\columnwidth,height=0.31\columnwidth]{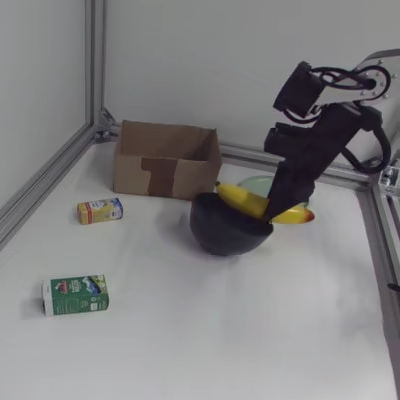}%
  }\hfill
  \subfloat[Push milk.]{%
    \includegraphics[width=0.31\columnwidth,height=0.31\columnwidth]{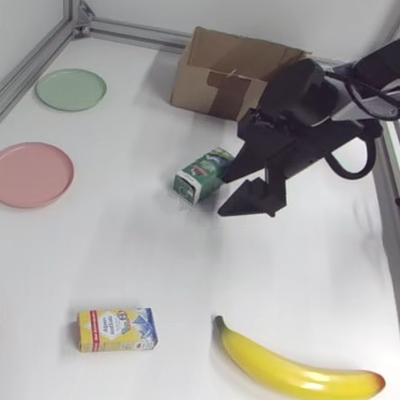}%
  }
  \caption{\textbf{Failure modes}. Models struggle with object grasping ($44\%$ of failure cases). The push task requires precision to align the gripper with the milk.}
  \label{fig:failure_modes}
\end{figure}

Adding human videos on average improves policy by $+20\%$ absolute across all tasks. Multi-view LAM training is a significant driver, improving on average by $+25\%$ over single-view ($46\%$ vs.\ $21\%$), explained by better action recoverability through the second view (\autoref{fig:realworld_probe_bars}) similar to our results in simulation. In \autoref{fig:realworld_probe}, we demonstrate latent transfer between human hands and robot rollouts.

'Novel placement' tasks achieve the highest SR (avg. $51\%$), followed by 'novel object' (avg. $34\%)$. Placement semantics do not require dexterous object interaction and can be carried over from human videos, leading to high SRs. In contrast, handling novel objects ($34\%$) and novel movements ($16\%$) requires object interaction that is difficult to recover without robot demonstrations.

While we observe generalization under human-only settings without any robot data, the SRs for these cases remain relatively low ($\approx 20\%$), limiting the practical applicability of zero-shot transfer. 

Scaling-wise, adding five robot demonstrations on the target embodiment lifts SR from $26\%$ to $42\%$ ($+16\%$) with video, and $6\%$ to $23\%$ ($+17\%$) for action-only. Even for high numbers of robot demonstrations for a task (50), adding human data improves policy success rate in real-world settings. This indicates that our video and robot data are complementary.

\textbf{Failure cases}. Analysis of failure cases shows that human videos mainly improve high-level planning. Adding videos reduces wrong-plan failures (e.g., gripping far off the object) from $52\%$ to $23\%$ of rollouts, while missed grips still account for $54\%$ of the remaining failures (see~\autoref{fig:failure_modes}). Simulation shows the same pattern, where grasping accounts for $51\%$ of failures on LIBERO. Grasping and object interaction remain a major failure case.

\section{Conclusion}

Our results show that spatial grounding through multi-view training is an effective and affordable lever for robot policy training. Calibrated dual-view datasets can surpass far larger single-view corpora, showing that simply scaling single-view data collection is not an effective way to recover 3D motions. Scaling embodiments is a secondary factor, and a single cross-embodiment such as a human hand is sufficient to achieve skill transfer. Video and robot demonstrations remain complementary; gains from video data are task-dependent and diminish on precision bottlenecks and tasks relying heavily on gripper actions, where more robot data supervision is required.

\renewcommand*{\bibfont}{\footnotesize}
\printbibliography

\end{document}